\documentclass[letterpaper]{article} 
\pdfoutput=1
\usepackage{aaai24}  
\usepackage{times}  
\usepackage{helvet}  
\usepackage{courier}  
\usepackage[hyphens]{url}  
\usepackage{graphicx} 
\usepackage{natbib}  
\usepackage{xcolor}
\usepackage{caption} 
\usepackage{algorithm}
\usepackage{algorithmic}

\usepackage{graphicx}
\usepackage{subcaption}

\usepackage{array}
\usepackage{diagbox}
\usepackage{multirow}
\usepackage{makecell}
\usepackage{booktabs}
\usepackage{amsmath} 
\usepackage{amsfonts} 
\usepackage{newfloat}
\usepackage{listings}

\DeclareCaptionStyle{ruled}{labelfont=normalfont,labelsep=colon,strut=off} 
\floatstyle{ruled}
\newfloat{listing}{tb}{lst}{}
\floatname{listing}{Listing}
\title{Spatial-Temporal Interplay in Human Mobility: A Hierarchical Reinforcement Learning Approach with Hypergraph Representation}

\author{
    Zhaofan Zhang\textsuperscript{\rm 1}\equalcontrib,
    Yanan Xiao\textsuperscript{\rm 2,5}\equalcontrib,
    Lu Jiang\textsuperscript{\rm 3},
    Dingqi Yang\textsuperscript{\rm 1,4},
    Minghao Yin\textsuperscript{\rm 2,5},
    Pengyang Wang\textsuperscript{\rm 1,4}\thanks{Corresponding author.}
}
\affiliations{
    \textsuperscript{\rm 1}The State Key Laboratory of Internet of Things for Smart City, University of Macau, China\\
    \textsuperscript{\rm 2}School of Computer Science and Information Technology, Northeast Normal University,  China\\
    \textsuperscript{\rm 3}Department of  Information Science and Technology, Dalian Maritime University, China\\
    \textsuperscript{\rm 4}Department of Computer and Information Science, University of Macau, China\\
    \textsuperscript{\rm 5}Key Laboratory of Applied Statistics of MOE, Northeast Normal University, China\\
    \{mc25684, dingqiyang, pywang\}@um.edu.mo,
    \{xiaoyn117, ymh\}@nenu.edu.cn,
    jiangl761@dlmu.edu.cn
}

\begin{document}
\maketitle

\begin{abstract}
In the realm of human mobility, the decision-making process for selecting the next-visit location is intricately influenced by a trade-off between spatial and temporal constraints, which are reflective of individual needs and preferences. 
This trade-off, however, varies across individuals, making the modeling of these spatial-temporal dynamics a formidable challenge. 
To address the problem, in this work, we introduce the ``Spatial-temporal Induced Hierarchical Reinforcement Learning'' (STI-HRL) framework, for capturing the interplay between spatial and temporal factors in human mobility decision-making. 
Specifically, STI-HRL employs a two-tiered decision-making process: the low-level focuses on disentangling spatial and temporal preferences using dedicated agents, while the high-level integrates these considerations to finalize the decision. 
To complement the hierarchical decision setting, we construct a hypergraph to organize historical data, encapsulating the multi-aspect semantics of human mobility. 
We propose a cross-channel hypergraph embedding module to learn the representations as the states to facilitate the decision-making cycle.
Our extensive experiments on two real-world datasets validate the superiority of STI-HRL over state-of-the-art methods in predicting users' next visits across various performance metrics.

\end{abstract}

\section{Introduction}

\begin{figure}[t]
\centering
\includegraphics[width=\linewidth]{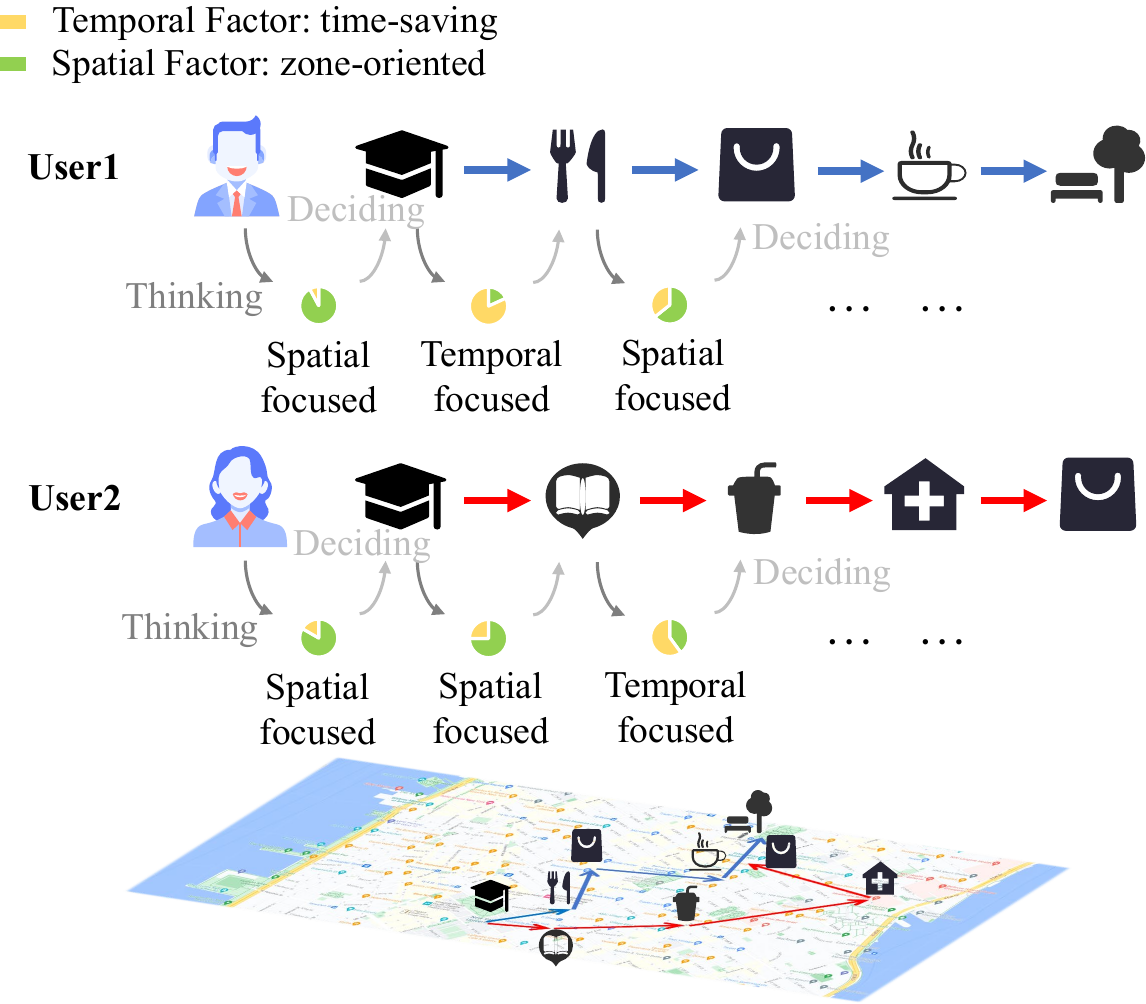} 
\caption{The two users with inconsistent spatial-temporal dynamics and the factors they prioritize when making decisions in various scenarios.}
\label{fig1}
\end{figure}

Human mobility refers to the patterns and behaviors associated with how and where individuals move or travel within physical spaces. 
Comprehending and accurately modeling human mobility offers valuable insights with applications spanning transportation~\cite{wang2020incremental}, urban planning~\cite{wang2018learning,wang2020reimagining}, public health~\cite{qureshi2021equitable}, business~\cite{antoniou2012human}, etc.
Specifically, individual preferences regarding spatial and temporal factors are pivotal in modeling human mobility. 
From a decision-making standpoint, the selection of the next-visit site by humans is deeply influenced by the interplay between spatial and temporal constraints, which echo individual needs and preferences. 
This spatial-temporal dynamic is subject to change over time and can differ among individuals. 
For instance, as illustrated in Figure \ref{fig1}, User 1 prioritizes temporal factors, opting for a quick dinner post-school. 
However, post-dinner, his preference shifts towards a superior shopping experience, emphasizing spatial considerations over time. 
In contrast, when comparing the mobility decisions of User 1 and User 2, their spatial-temporal dynamics differ significantly.

%
Existing works on studying on spatial-temporal factors in human mobility fall into two categories: 
(1) Temporal-oriented approaches primarily emphasize the time dimension, modeling preferences that evolve over time, while incorporating spatial context as a secondary element. 
For instance, RNN-based methods~\cite{DBLP:conf/www/WangYCHWZH20,zhao2020go} employ transition matrices or gates to merge spatial context with temporal factors within RNNs' recurrent hidden states. 
Some models~\cite{lian2020geography,DBLP:conf/www/LuoLL21} further leverage attention mechanisms to enhance spatial-temporal integration. 
(2) Spatial-oriented strategies prioritize the spatial dimension, preserving spatial autocorrelation through graph representation learning and viewing the interplay as temporal graph dynamics. 
Recent studies harness high-order topological structures for graph representations and capture dynamics using sequential models~\cite{wang2019adversarial,yang2019revisiting,wang2020exploiting,wang2022learning,wang2023reinforced}. 
Despite promising results achieved, a common limitation of these methods is their predominant reliance on either the spatial or temporal dimension, potentially oversimplifying the intricate interactions between the two in decision-making.
This presents a pressing research challenge: how can we effectively capture the spatial-temporal interplay inherent in mobility decision-making?

To address the challenge, we propose ``Spatial-Temporal Induced Hierarchical Reinforcement Learning'' (STI-HRL) framework, which formulates the human mobility as a two-layered decision-making process with a spatial-temporal Decoupling-Integration schema. 
Specifically, STI-HRL is built following the hierarchical setting
(1) the low-level, that decouples spatial-temporal interplay with a spatial agent and a temporal agent to focus on each dimension, respectively; 
(2) the high-level, that integrates the insights from two agents in the low-level to make the final decision on mobility. 
To facilitate STI-HRL with a proper environment, we propose a Mobility Hypergraph to organize the multi-aspect semantics of mobility records. 
The hyperedge embeddings from this Mobility Hypergraph serve as the state to support the decision-making process in STI-HRL.

Our contributions can be summarized as follows:

\begin{itemize}
\item We introduce a novel formulation of human mobility modeling, conceptualizing it as a two-tiered decision-making process that emphasizes the spatial-temporal interplay in mobility decisions.
\item To instantiate this formulation, we develop STI-HRL, offering a unique pipeline for modeling human mobility decisions.
\item To preserve the multi-faceted semantics of mobility records, we devise the Mobility Hypergraph and introduce an effective hypergraph embedding method to represent the environment's state.
\item We conduct extensive experiments to validate the effectiveness of our proposed STI-HRL. The results demonstrate the superior performance of STI-HRL, consistently outperforming baseline algorithms.
\end{itemize}

\section{Definitions and Problem Formulation}

\begin{figure}[!t]
\centering
\includegraphics[width=\linewidth]{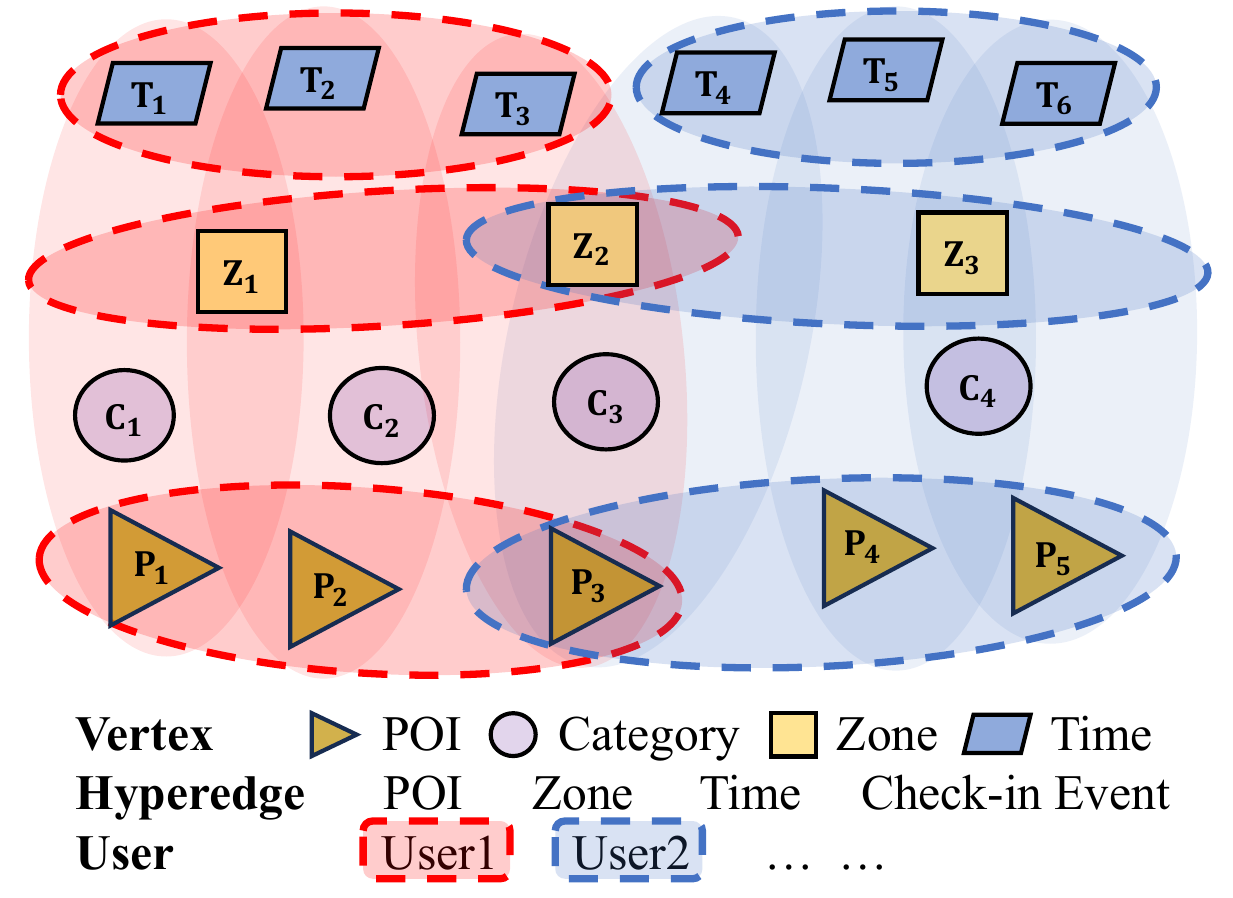}
\caption{A toy example of Mobility Hypergraph.}
\label{fig:mobility hypergraph}
\end{figure}
\subsection{Mobility Hypergraph}
In order to capture the complicated spatial-temporal interplay of human mobilities, we propose to construct a hypergraph to represent historical records, called Mobility Hypergraph. 
Formally, the Mobility Hypergraph $\mathcal{G}$ is defined as $\mathcal{G} = (\mathbf{V}, \mathbf{E})$, where $\mathbf{V}$ and $\mathbf{E}$ denote the vertex set and hyperedge set, respectively. 

\noindent \textbf{Vertices}. 
Mobility Hypergraph aims to organize the check-in events with preserving multi-aspect semantics. 
Specifically, we decompose the check-in events records $\mathbb{H}$ into four semantic channels, including (1) POI channel, denoted as $\mathbf{P}$; (2) POI category channel, denoted as $\mathbf{C}$; (3) zone channel, denoted $\mathbf{Z}$; and (4) time channel, denoted as $\mathbf{T}$. 
In this work, we consider four types of vertices in corresponding to the four semantic channels. 
Then, the vertex set can be denoted as $\mathbf{V}= \mathbf{P} \cup \mathbf{C} \cup \mathbf{Z} \cup  \mathbf{T}$.

\noindent \textbf{Hyperedges}. 
A hyperedge connects two or more vertices. We identify four types of hyperedges:
(1) POI hyperedge, linking all POI vertices a user has visited;
(2) zone hyperedge, connecting all the zone vertices visited by a user;
(3) time hyperedge, associating all the time vertices visited by a user;
(4) event hyperedge, which interlinks the POI, category, zone, and time vertices of a specific check-in event.

Figure~\ref{fig:mobility hypergraph} provides a toy example of the Mobility Hypergraph for two users. A user's historical records are structured into POI, zone, and time hyperedges, along with all the event hyperedges they have visited. Notably, while the first three hyperedge types are homogeneous (connecting vertices of the same semantic channel), the event hyperedges are heterogeneous, linking vertices across all semantic channels. This structure not only encapsulates the spatial-temporal nuances of each aspect but also establishes connections between users.

\begin{figure*}[!th]
  \centering
\includegraphics[width=1\linewidth]{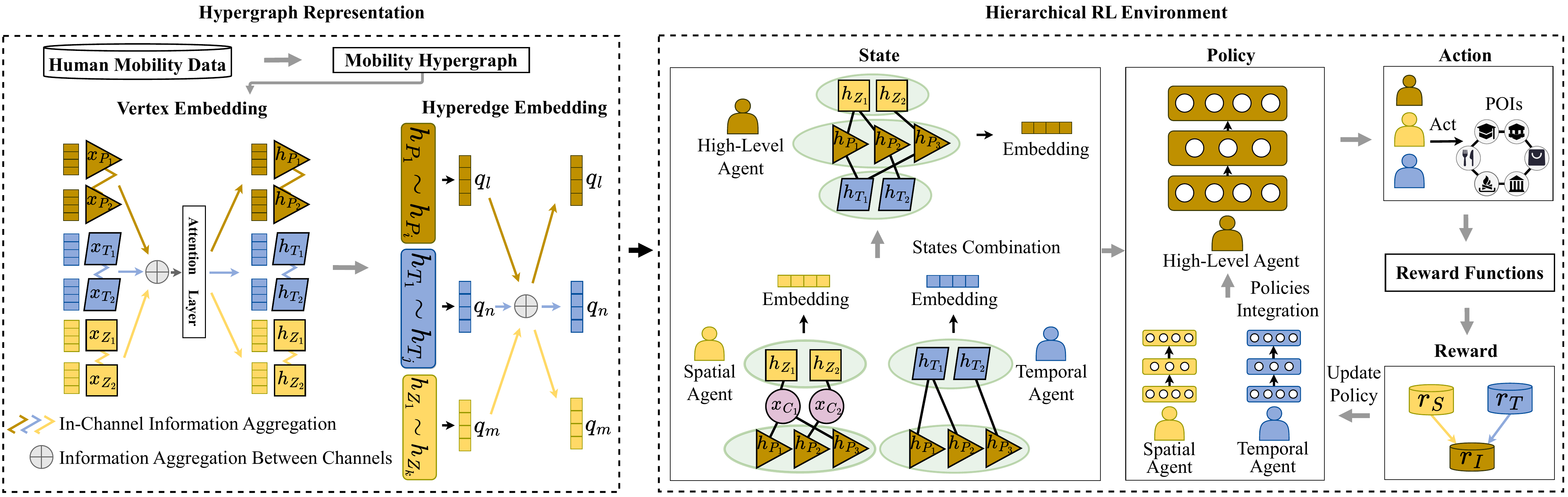}
  \caption{Spatial-Temporal Induced Hierarchical Reinforcement Learning.}
  \label{fig:Loss}
\end{figure*}

\subsection{Problem Formulation}

In this work, we formulate human mobility as a Markov Decision Process (MDP) that users make decisions on the next-visit location based on historical records reflecting individual preferences in certain spatial-temporal contexts. The key components of human mobility MDP are defined as
(1) \textbf{States} $S$. Each state $\mathbf{s} \in S$ represents a specific spatial-temporal context derived from historical records, which are organized as a Mobility Hypergraph. 
(2) \textbf{Actions} $A$. Each action $a \in A$ corresponds to a potential next-visit location. 
(3) \textbf{Transition Probabilities} $\Gamma$. $\Gamma(\mathbf{s}'|\mathbf{s}, a)$ represents the probability of transitioning from state $\mathbf{s}$ to state $\mathbf{s}'$ when action $a$ is taken. This probability can be estimated from historical records, reflecting how often a user transitions from one spatial-temporal context to another after choosing a particular location.
(4) \textbf{Rewards} $R$. $R(\mathbf{s}, a, \mathbf{s}')$ denotes the reward received after transitioning from state $\mathbf{s}$ to state $\mathbf{s}'$ due to action $a$. The reward can be designed to reflect user satisfaction or any other metric of interest. We will introduce the reward design later.
(5) \textbf{Environment} $E$. The environment consists of all participants of check-in events. It responds to the user's action by providing a new state and a reward. The environment's dynamics are governed by the transition probabilities $\Gamma$ and the reward function $R$.
(6) \textbf{Policy} $\pi$. A policy $\pi$ defines how users take action. Specifically, $\pi(\mathbf{s})$ gives the probability distribution over actions in state $\mathbf{s}$. The goal in the MDP is to find an optimal policy $\pi^*$ that maximizes the expected cumulative reward over time.

Then, the human mobility MDP can be represented as a tuple:
$(S, A, \Gamma, R, E) $.
Given the human mobility MDP formulation, we aim to develop a hierarchical reinforcement learning framework find the optimal policy $\pi^*$ that guides users in making decisions on the next-visit location with considering spatial-temporal interplay.

\section{Hypergraph Embedding for State Representation}
In our study, human mobility records are systematically structured into a Mobility Hypergraph. 
Therefore, we employ hypergraph embeddings to represent the state to facilitate the hierarchical reinforcement learning framework. 
The process begins by learning vertex embeddings. 
Subsequently, these vertex embeddings are aggregated to form hyperedge embeddings. 
Along this line, the embeddings effectively capture the multi-aspect semantics of human mobilities, encompassing both spatial and temporal contexts. 
Next, we introduce the embedding procedure in detail.

\noindent\textbf{Vertex Embedding}.   
We denote the raw features of vertex $v_i \in \mathcal{V}$ as $\mathbf{x}_i \in \mathbb{R}^d$, and $\mathcal{N}_i$ represents vertex $v_i$'s neighbors that are within the hyperedges. 
We employ the attention mechanism to capture the interrelationship between vertices and the respective neighbors in the same channel. 
Specifically, for the vertex $v_i$ and its neighbor $v_j$ ($j \in \mathcal{N}_i$), the attention coefficient $\alpha_{ij}$ can be represented as

\begin{equation}
\alpha_{ij} = \frac{\exp(\sigma(\mathbf{W}_{\alpha}^T[\mathbf{W}_v \mathbf{x}_i \| \mathbf{W}_v \mathbf{x}_j]))}{\sum_{j \in \{\mathcal{N}_i, i \}}\exp(\sigma( \mathbf{W}_{\alpha}^T[\mathbf{W}_v \mathbf{x}_i \| \mathbf{W}_v \mathbf{x}_j]))}, 
\end{equation}
where $\sigma$ denotes the activation function,  $\mathbf{W}_{\alpha}$ denotes the weight matrix for attention calculation, and $\mathbf{W}_v$ denotes the learnable weight matrix for feature transformation. 
Then, the embedding $\mathbf{h}_i$ of the vertex $v_i$ can be represented by aggregating the neighbors' information as
\begin{equation}
\mathbf{h}_i = \sigma \left( \sum_{j \in \{\mathcal{N}_i, i   \} } \alpha_{ij} \mathbf{W}_v \mathbf{x}_j \right).
\end{equation}

\noindent\textbf{Hyperedge Embedding}. 
We define four types of hyperedges, 
among which, POI, zone, and time hyperedges are homogeneous, while event hyperedges are heterogeneous. 
Specifically, for the homogeneous hyperedges $e_i \in \mathbf{E}$, we first aggregate all the vertex embedding within the hyperedge to represent the hyperedge embedding. 
The hyperedge embedding  $\mathbf{q}_i$ can be denoted as 
    \begin{equation}
    \mathbf{q}_i = \sigma \left(\sum_{j \in |e_i|} \mathbf{h}_j \right), 
    \end{equation}
where $|e_i|$ denotes all the associated vertices in $e_i$.

The event hyperedges serve as a bridge to link the semantics from different channels flowing through the hypergraph topology. 
We then update the hyperedge embedding $\mathbf{q}_i$ by aggregating information from hyperedges on other channels that are interlinked by the same event hyperedge:
\begin{equation}
    \mathbf{q}_i = \sigma \left( \sum_{k \in \Phi(e_i)} \mathbf{W}_{\Psi(e_k)} \mathbf{q}_k \right),
    \end{equation}
where $\Phi(e_i)$ denotes the query function that retrieves hyperedges from alternate channels that are interconnected by the same event hyperedge as the given hyperedge, $\Psi(\cdot)$ is the function to return the type of the given hyperedge,  and $\mathbf{W}_{\Psi(e_k)}$ denotes the aggregation weights for the given the type $\Psi(e_k)$.

The homogeneous hyperedge embedding $\mathbf{q}$ captures users' preferences from various channels across both spatial and temporal dimensions, which will be taken as the state to represent the status of the environment.
In the next section, we detail the hierarchical reinforcement learning setting of our proposed STI-HRL. 
We explain how it integrates insights from the hypergraph representation to make optimized mobility decisions. 

\section{Spatial-Temporal Induced Hierarchical Reinforcement Learning}

\subsection{Low-level: Decoupling Spatial-Temporal Interplay}
As the first phase of STI-HRL, the low level seeks to distinguish users' preferences within the spatial and temporal dimensions. 
STI-HRL employs a dual-agent setting, comprising a spatial agent and a temporal agent, to discern human mobility decision patterns within these dimensions, respectively. The subsequent sections detail the components and mechanics of the low-level decision-making in STI-HRL.

\noindent\textbf{States.} 
We take the hyperedge embeddings as the representation of states. 
Specifically, for the spatial agent, the state aims to capture the preferences and interactions on spatial factors. 
Therefore, we concatenate the associated POI hyperedge embeddings and zone hyperedge embeddings to represent the state. Formally, let $\mathbf{s}^{u}_{S,t}$ represent the spatial agent state for a user $u$ at time $t$,

\begin{align}
\begin{split}
    &\mathbf{s}^{u}_{S,t} = \text{CONCATENATE} (\mathbf{q}_l, \mathbf{q}_m) \\
    &\text{s.t., }l=\Theta_{\mathbf{P}}(u) \text{ \& }
 m=\Theta_{\mathbf{Z}}(u), 
 \end{split}
\end{align}
where $\Theta_{\mathbf{P}}(u)$ and $\Theta_{\mathbf{Z}}(u)$ denote the indexes of associated POI hyperedge and zone hyperedge for the user $u$, respectively. 
Similarly, for the temporal agent, we concatenate the associated POI hyperedge embeddings and time hyperedge embeddings to represent the temporal agent state as 

\begin{align}
\begin{split}
    &\mathbf{s}^{u}_{T, t} = \text{CONCATENATE} (\mathbf{q}_l, \mathbf{q}_n) \\
    &\text{s.t., }l=\Theta_{\mathbf{P}}(u) \text{ \& }
 n=\Theta_{\mathbf{T}}(u), 
 \end{split}
\end{align}
where $\Theta_{\mathbf{T}}(u)$ denotes the index of the associated time hyperedge for the user $u$.

\noindent\textbf{Reward Design. } 
For the spatial agent, we aim to evaluate the decision-making performance in terms of the following three aspects: 
(1) POI-POI geographic distance, denoted as $ r_d $, which is the reciprocal of the geographic distance between the actual and predicted next-visited POI; 
(2) POI-POI category similarity, denoted as $ r_c $, which is the cosine similarity between predicted and actual POI categories (computed using fastText embeddings~\cite{bojanowski2017enriching}); and 
(3) POI-POI relative ranking $ r_p $, which is the ranking order of the actual visited POI in the predicted POI candidate list. 
Then, we define the reward for the spatial agent as the combination of the above three components as:
\begin{equation}
r_{S} = w_{d} \cdot r_{d} + w_{c} \cdot r_{c} + w_{ps} \cdot r_{p}, 
\end{equation}
where  $ w_{d} $, $ w_{c} $, and $ w_{ps} $ denote the weights for balancing the influence of $r_d$, $r_c$, $r_p$, respectively. 

While for the temporal agent, we focus more on the time factors of mobility. 
In addition to the POI-POI relative ranking $ r_p $ mentioned above, we also include another measurement $r_t$, the Kullback–Leibler (KL) divergence between the visiting frequency distribution of the actual and predicted POIs across the time span. 
Then, we combine the two measurements to define the reward for the temporal agent as
\begin{equation}
    r_{T} = w_{t} \cdot r_{t} + w_{pt} \cdot r_{p}, 
\end{equation}
where $w_{t}$ and $w_{pt}$ denote the weights for balancing the influence of $r_t$ and $r_p$, respectively.

\subsection{High-level: Integrative Synthesis of Space-Time Dynamics} 
The high-level decision process employs a dedicated agent to integrate insights from the low-level agents, synthesizing spatial and temporal considerations in interaction with the Mobility Hypergraph to produce the final human mobility decision. 
In this section, we introduce the key components of the high-level agent, and outline the workflow.

\noindent\textbf{State.} 
To encapsulate the interplay of spatial and temporal dynamics, the state of the high-level process is defined by combining the states of the two low-level agents.
Still taking the state for user $u$ at time $t$ for illustration, the state of the high-level integrative process $\mathbf{s}_{I,t}^{u}$ can be represented as 
\begin{equation}
    \mathbf{s}_{I,t}^{u} = \lambda_{T} \cdot \mathbf{s}_{T,t}^{u} + \lambda_{S} \cdot \mathbf{s}_{S,t}^{u},
\label{equ: high level state}
\end{equation}
where $\lambda_S$ and $\lambda_T$ denote the weights for the spatial state $\mathbf{s}_{S,t}^{u}$ and temporal state $\mathbf{s}_{T,t}^{u}$, respectively.

\noindent\textbf{Reward.} 
Similarly, in terms of the high-level reward, we calibrate the interplay between spatial and temporal realms by amalgamating the rewards from the low-level agents. 
Specifically, let $w_{S}$ and $w_{T}$ denote the weights for spatial and temporal rewards, respectively. 
Then, the reward for the high-level agent $r_{I}^{u}$ can be represented as:
\begin{equation}
    r^{u}_I = w_{T} \cdot r_{T}^{u} + w_{S} \cdot r_{S}^{u}.
\label{equ: high-level reward}
\end{equation}

\begin{table*}[htbp]
\centering
\small
\newcolumntype{C}[1]{>{\centering\arraybackslash}p{#1}}

\begin{tabular}{c|C{1.8cm}|p{0.8cm}p{0.8cm}p{0.8cm}|p{0.8cm}p{0.8cm}p{0.8cm}|p{0.8cm}p{0.8cm}p{0.8cm}|p{0.8cm}p{0.8cm}p{0.8cm}}

\toprule

& \multirow{2}{*}{\diagbox[width=2cm]
{\makebox[25pt][r]{\textbf{Method}}}
{\qquad\smallskip\textbf{Metrics}}} 
&\multicolumn{3}{c|}{\textbf{Recall}}
&\multicolumn{3}{c|}{\textbf{F1}}
&\multicolumn{3}{c|}{\textbf{MRR}}
&\multicolumn{3}{c}{\textbf{NDCG}}
\\
\cmidrule(lr){3-14} 
&  
& \textbf{@5} & \textbf{@10} & \textbf{@20} 
& \textbf{@5} & \textbf{@10} & \textbf{@20} 
& \textbf{@5} & \textbf{@10} & \textbf{@20}
& \textbf{@5} & \textbf{@10} & \textbf{@20}\\
\midrule
\multirow{9}{*}{\rotatebox[origin=c]{90}{\textbf{New York}}}

&ST-RNN & 0.0037 & 0.0075 & 0.0121 & 0.0012 & 0.0013 & 0.0011 & 0.0017 & 0.0022& 0.0025 & 0.0022 & 0.0037& 0.0045\\
& DeepMove & 0.3857 & 0.4724 & 0.5462 & 0.1285 & 0.0859 & 0.0520 & 0.2501 & 0.2618& 0.2670 & 0.2839 & 0.3120 & 0.3308\\
&  LSTPM & 0.4039 & 0.5064 & 0.5958 & 0.1346 & 0.0920 & 0.0567 & 0.2698 & 0.2836 &0.2989 & 0.3232 & 0.3364 & 0.3591 \\
& HST-LSTM & 0.3158 & 0.3795 & 0.4345 & 0.1025 & 0.0690 & 0.0413 & 0.2139 & 0.2225& 0.2264 & 0.2393 & 0.2600& 0.2740\\
&  STAN & 0.2754 & 0.3742 & 0.4688 & 0.0918 & 0.0680 & 0.0446 & 0.1619 & 0.1751& 0.1817 & 0.1900 & 0.2220& 0.2460 \\
& LBSN2vec & 0.3559 & 0.4250 & 0.4544  & 0.0846  &  0.0622 &  0.0408 & 0.2041  & 0.2282 & 0.2551  &  0.2599 & 0.2815 & 0.3122\\
& FPMC & 0.2556 & 0.3437 & 0.4361 & 0.0852 & 0.0624 & 0.0415 & 0.1524 & 0.1642& 0.1706 & 0.1780 & 0.2064& 0.2298\\
& KNN Bandit & 0.2045 & 0.2552 & 0.2778  & 0.1414  &  0.0717 & 0.0466  & 0.1644  & 0.1944 &  0.2478 &  0.2522 & 0.2841  & 0.3015 \\
& \textbf{STI-HRL} & 
\textbf{0.4268} & \textbf{0.5227} & \textbf{0.6298}  & \textbf{0.1467}  & \textbf{0.1104}  & \textbf{0.0638}  & \textbf{0.2930}  & \textbf{0.3066} & \textbf{0.3275}  & \textbf{0.3444}  & \textbf{0.3685 }& \textbf{0.4018}  \\

\midrule
&  \textbf{Improvement} & 
\textbf{5.67\%}& \textbf{3.22\%}& \textbf{5.71\%}& 
\textbf{8.90\%}& \textbf{13.5\%}& \textbf{12.52\%}& 
\textbf{8.60\%}& \textbf{8.11\%}& \textbf{9.57\%}& 
\textbf{6.56\%}& \textbf{8.71\%}& \textbf{10.36\%}\\

\midrule
\multirow{9}{*}{\rotatebox[origin=c]{90}{\textbf{Tokyo}}}

&ST-RNN & 0.0044 & 0.0068 & 0.0085 & 0.0014 & 0.0012 & 0.0008 & 0.0021 & 0.0024& 0.0025 & 0.0027 & 0.0034 & 0.0039\\
& DeepMove& 0.4124 & 0.4879 & 0.5510 & 0.1374 & 0.0887 & 0.0524 & 0.2832 & 0.2935& 0.2979 & 0.3155 & 0.3401& 0.3561 \\
&  LSTPM & 0.4046 & 0.4864 & 0.5614 & 0.1348 & 0.0884 & 0.0534 & 0.2789 & 0.2899& 0.2951 & 0.3102 & 0.3367 & 03557\\
& HST-LSTM & 0.3090 & 0.3783 & 0.4462 & 0.1030 & 0.0687 & 0.0424 & 0.2137 & 0.2229& 0.2276 & 0.2374 & 0.2598& 0.2769\\
&  STAN& 0.1990 & 0.2574 & 0.3196 & 0.0663 & 0.0468 & 0.0304 & 0.1300 & 0.1378& 0.1421 & 0.1471 & 0.1660& 0.1817 \\
& LBSN2vec & 0.3551 & 0.4122 & 0.4753  & 0.0944  &  0.0713 &  0.0412 & 0.2274  & 0.2589 & 0.2637 &  0.2476 & 0.2664 & 0.3067\\
& FPMC & 0.1086 & 0.1557 & 0.2209 & 0.0362 & 0.0283 & 0.0210 & 0.0623 & 0.0684& 0.0729& 0.0737 & 0.0888 & 0.1052\\
& KNN Bandit & 0.2388 & 0.2842 & 0.3104  & 0.0947  &  0.0764 &  0.0573 & 0.1441  & 0.1951 &  0.2174 & 0.2006  & 0.2504 & 0.2894  \\
& \textbf{STI-HRL}  
& \textbf{0.4425} &\textbf{0.5150}  & \textbf{0.5904}  & \textbf{0.1552}   & \textbf{0.1005}  &  \textbf{0.0608} & \textbf{0.3031}  & \textbf{0.3277} & \textbf{0.3349}  &  \textbf{0.3361 }& \textbf{0.3615} & \textbf{0.3991} \\

\midrule
&  \textbf{Improvement} & 
\textbf{7.30\%}& \textbf{5.55\%}& \textbf{5.17\%}& 
\textbf{13.10\%}& \textbf{13.30\%}& \textbf{13.86\%}& 
\textbf{7.03\%}& \textbf{11.65\%}& \textbf{12.42\%}& \textbf{6.53\%}& \textbf{6.29\%}& \textbf{12.08\%}\\

\bottomrule
\end{tabular}
\caption{Overall performance comparison. The best performance for each metric is highlighted in bold. The ``Improvement'' section denotes the percentage increase over the second-best models.}
\label{Experment_Table}
\end{table*}

\subsection{Policy Learning} 

\noindent \textbf{Workflow. }
In the low-level process of STI-HRL, 
the spatial agent, using hyperedge embeddings, derives its state $ \mathbf{s}^{u}_{S,t} $ to predict the next location a user might visit, receiving feedback through the reward $ r_S $. 
Similarly, the temporal agent uses embeddings to model its state $ \mathbf{s}^{u}_{T,t} $, aiming to predict when the visit event occurs, with its efficacy gauged by $ r_T $. Their states transit as:

\begin{equation}
\mathbf{s}^{u}_{\triangleright,t+1} = \Gamma(\mathbf{s}^{u}_{\triangleright,t}, a_{t}, r_\triangleright), 
\end{equation}
where ``$\triangleright$'' denotes $S$ or $T$, for the spatial or temporal agent, respectively.
The high-level agent then integrates insights from the spatial and temporal agents. It synthesizes a holistic representation of user behavior and transitions as:

\begin{equation}
 \mathbf{s}^{u}_{I,t+1} = \Gamma(\mathbf{s}^{u}_{I,t}, a_{t}, r_{I}),
\end{equation}

\noindent \textbf{Optimization. }
We implement Multi-layer Perceptrons (MLP) for spatial, temporal and high-level policies, parameterized by $\theta_{S}$, $\theta_{T}$, and $\theta_{I}$, respectively. 
To make a succinct and unified description, we take ''$*$'' to represent the spatial $S$, temporal $T$, or the high-level integration $I$ for the following illustration.
The STI-HRL method employs policy gradient \cite{agarwal2020optimality} methods. 
 
The primary objective is to maximize the expected rewards:

\begin{equation}
    J(\theta_*) = \mathbb{E}[R_{*} | \theta_{*}],
\end{equation}
where $ J(\theta_*) $ is the expected reward under policy parameters $ \theta_* $ and $ R_* $ represents the reward of the agents. To achieve this, agent policies are refined by:

\begin{equation}
      \theta_{*}^{new} = \theta_{*}^{old} + \nabla_{\theta_{*}} J(\theta_*),
\end{equation}

\noindent \textbf{Policy Integration and Adaptation.} The high-level agent seamlessly integrates policies from the low-level agents:

\begin{align}
 \pi_{I}(a|\mathbf{s}_{I};\theta_{I}) &= \beta \pi_S(a|\mathbf{s}_S;\theta_S) \nonumber \\
 &+ (1-\beta) \pi_T(a|\mathbf{s}_T; \theta_T),
\end{align}
where $ \beta $ is the weight to determine the balance between spatial and temporal inputs. Recognizing the ever-evolving nature of spatial-temporal dynamics, the high-level agent dynamically adjusts $ \beta $ based on the relative performance of the integrated policies:

\begin{equation}
    \beta^{new} = \beta^{old} + \eta \nabla J(\theta_{I}),
    \label{equ: high level policy}
\end{equation}
where $ \eta $ represents a specific learning rate for weight adjustments. 
Moreover, the weights $\lambda_S, \lambda_T, w_T, w_S$ for the high-level agent in Equation~\ref{equ: high level state} and Equation~\ref{equ: high-level reward} also follow the same updating strategy with $\beta$ in Equation~\ref{equ: high level policy}.

\section{Experiments}
We evaluate the performance of STI-HRL in the next-location prediction task that is to predict the next location to visit given the historical mobility records. 
We aim to answer the following five primary research questions:

\begin{itemize}
    \item \textbf{Q1:} How well does STI-HRL perform in the next-location prediction task?
    \item \textbf{Q2:} How significant are the roles of spatial and temporal factors in mobility decision-making, respectively? 
    \item \textbf{Q3:} Is the design of STI-HRL reasonable in investigating spatial-temporal interplay? 
    \item \textbf{Q4:} Is it necessary and reasonable to construct Mobility Hypergraph for organizing human mobility records?
    \item \textbf{Q5:} How does the reward design  of each agent affect the performance of STI-HRL? 
\end{itemize}

\subsection{Experiment Settings}
\noindent \textbf{Datasets.} 
We conduct the experiment on two widely-used check-in datasets of New York (NYC) and Tokyo (TKY) from April 2012 to February 2013~\cite{yang2014modeling}.
Each entry in the dataset details user ID, POI ID, category, GPS coordinates, and the timestamp of each check-in event. 
In the experiment, we take the initial $70\%$ serving as the training set, the subsequent $10\%$ for validation, and the final $20\%$ for testing. 

\begin{figure}[!th]
\centering
\begin{subfigure}{.24\textwidth}
  \centering
  \includegraphics[width=\linewidth]{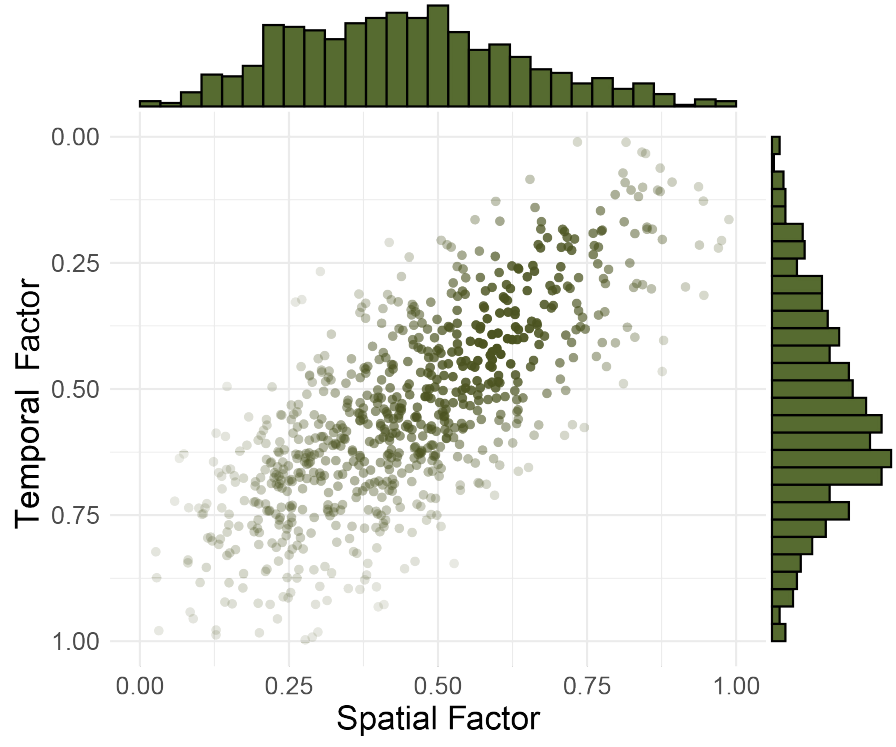}
  \scriptsize (a) New York
\end{subfigure}%
\begin{subfigure}{.24\textwidth}
  \centering
  \includegraphics[width=\linewidth]{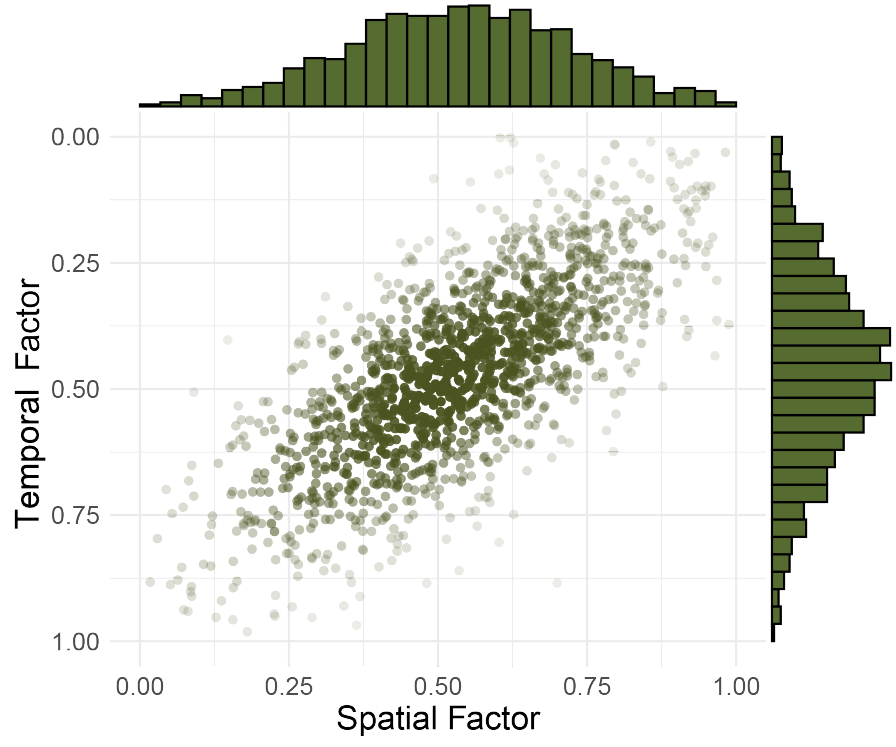}
  \scriptsize (b) Tokyo
\end{subfigure}
\caption{A visualization of the weight distribution for spatial ($\beta$) and temporal ($1-\beta$) factors in the final decision.}
\label{fig: spatial-temporal factors}
\end{figure}

\begin{figure}[!th]
\centering
\begin{subfigure}{.24\textwidth}
  \centering
  \includegraphics[width=\linewidth]{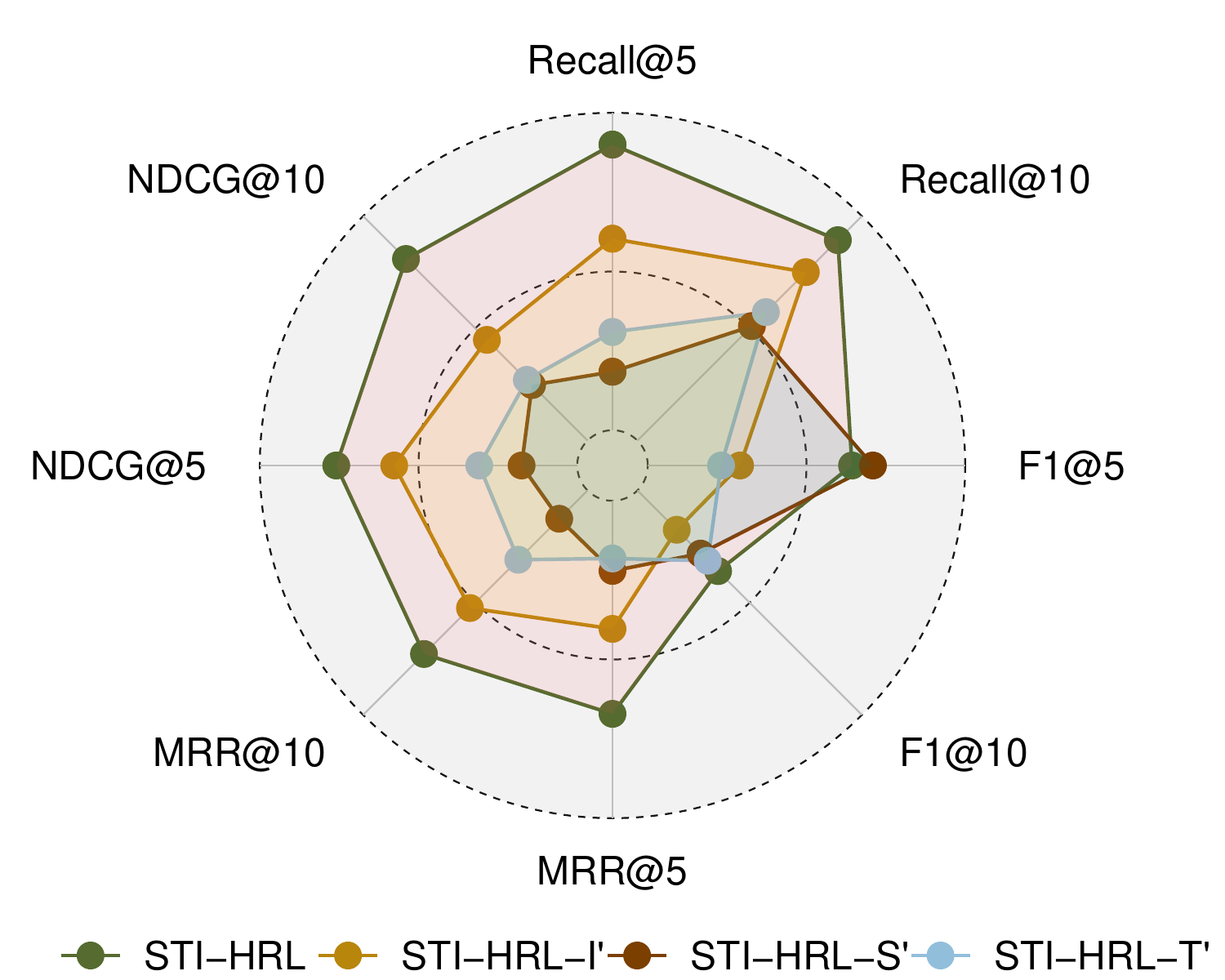}
  \scriptsize (a) New York
\end{subfigure}%
\begin{subfigure}{.24\textwidth}
  \centering
  \includegraphics[width=\linewidth]{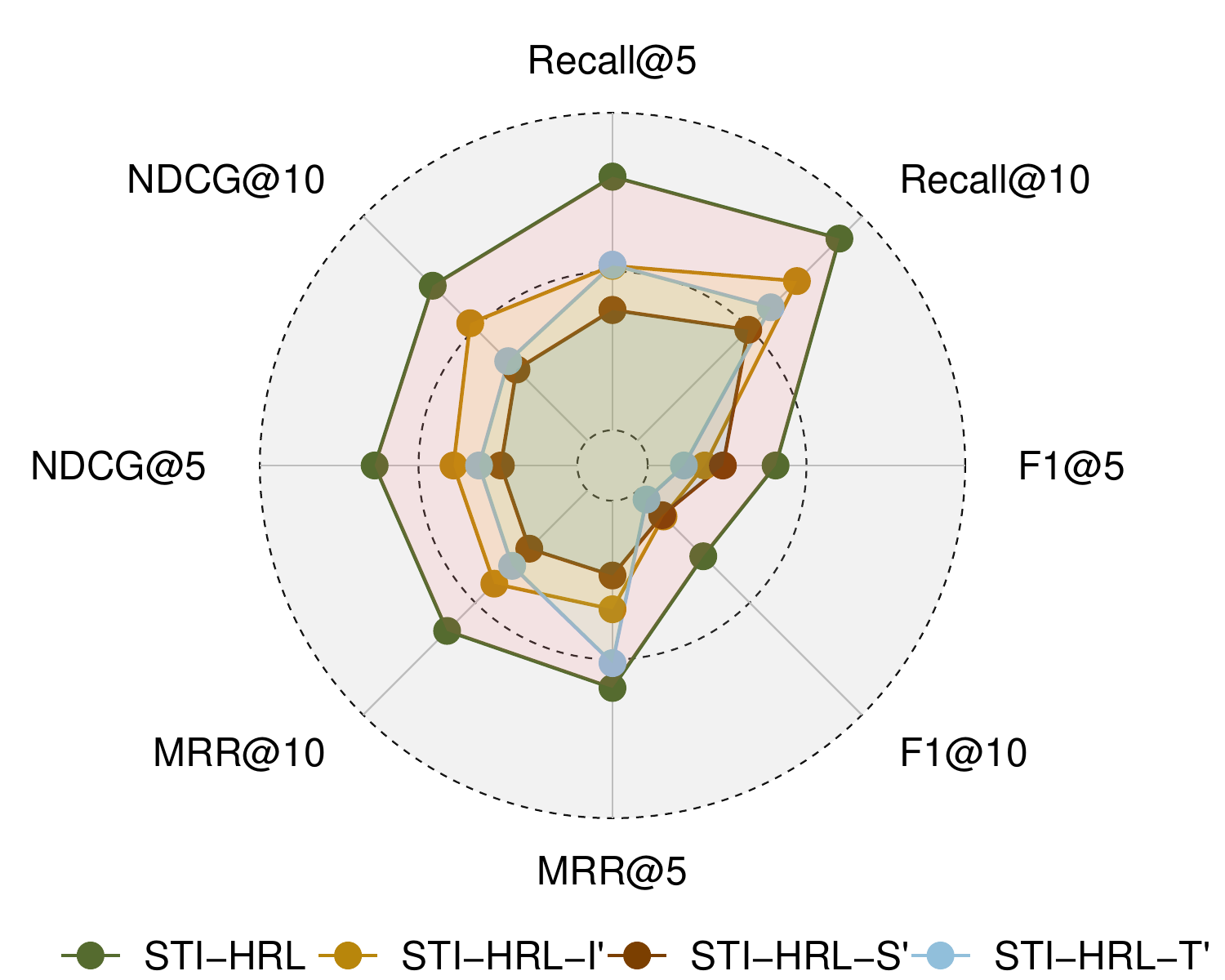}
  \scriptsize (b) Tokyo
\end{subfigure}
\caption{An ablation study of STI-HRL agent architecture. STI-HRL-$S'$, STI-HRL-$T'$, STI-HRL-$I'$ denote the variants of STL-HRL that spatial, temporal, and high-level integration removed agents, respectively.}
\label{Fig: ablation study on agents}
\end{figure}

\noindent \textbf{Baseline Methods.} 
We compare STI-HRL with eight baseline algorithms, including (1) \textbf{ST-RNN}~\cite{liu2016predicting}, (2) \textbf{DeepMove}~\cite{feng2018deepmove}, (3) \textbf{LSTPM}~\cite{sun2020go}, (4) \textbf{HST-LSTM}~\cite{DBLP:conf/ijcai/Kong018}, (5) \textbf{STAN}~\cite{DBLP:conf/www/LuoLL21}, (6) \textbf{LBSN2Vec}~\cite{yang2019revisiting}, (7) \textbf{FPMC}~\cite{rendle2010factorizing}, and (8) \textbf{KNN Bandit}~\cite{sanz2019simple}. 
Detailed descriptions can be found in Appendix A.

\noindent \textbf {Hyperparameter Settings.} 
We set the hypergraph embedding dimension as 64, and employ 4 attention heads. 
We utilize Adam optimizer~\cite{diederik2014adam} and set the learning rate as $0.001$ for optimization. 
Due to the page limitation, we only present the key hyperparameters here. 
The full description of hyperparameter settings can be found in Appendix B.

\subsection{Overall Performance (Q1)}
We present the overall performance comparison in Table~\ref{Experment_Table}. 
The results indicate that our proposed STI-HRL consistently outperforms all the baselines on both datasets. 
Specifically, for the New York dataset, STI-HRL can overpass the second-best baseline by more than $5.6\%$ on Recall, $8\%$ on F1, $8.6\%$ on MRR, and $6\%$ on NDCG; 
for the Tokyo dataset, the improvement is more significant with more than $7\%$ on Recall, $13\%$ on F1, $7\%$ on MRR, and $6.5\%$ on NDCG.
Such consistent improvement underscores the significance of spatial-temporal interplay in human mobility, and validates the effectiveness of our proposed STI-HRL.

\begin{figure}[!th]
\centering
\begin{subfigure}{.24\textwidth}
  \centering
  \includegraphics[width=\linewidth]{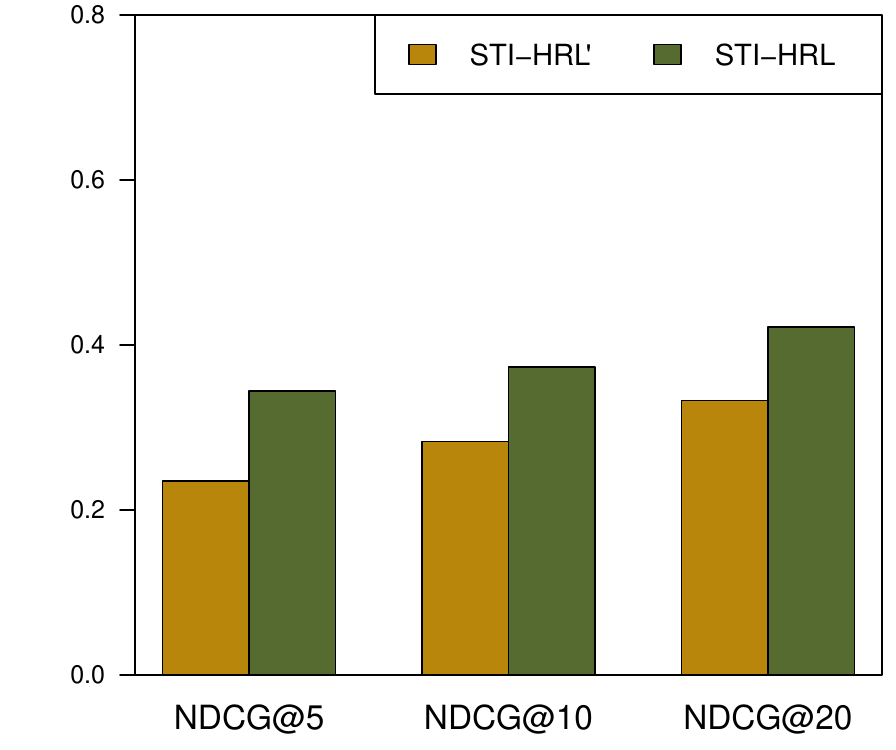}
  \scriptsize (a) New York
\end{subfigure}%
\begin{subfigure}{.24\textwidth}
  \centering
  \includegraphics[width=\linewidth]{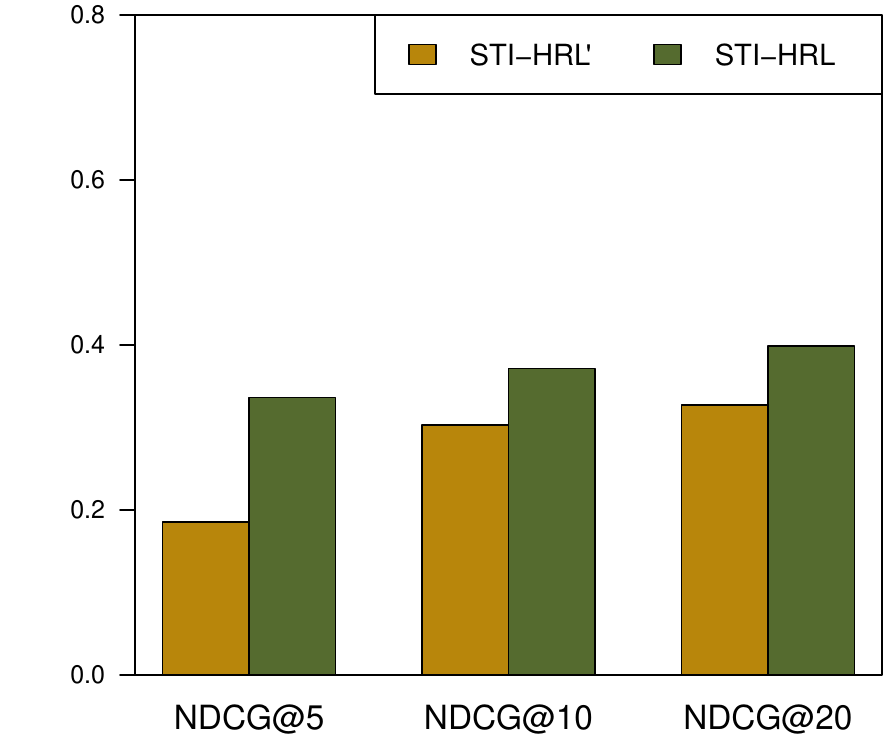}
  \scriptsize (b) Tokyo
\end{subfigure}
\caption{An ablation study on hypergraph.}
\label{fig:ablation study on hypergarph}
\end{figure}

\begin{figure}[!th]
\centering
\begin{subfigure}{.24\textwidth}
  \centering
  \includegraphics[width=\linewidth]{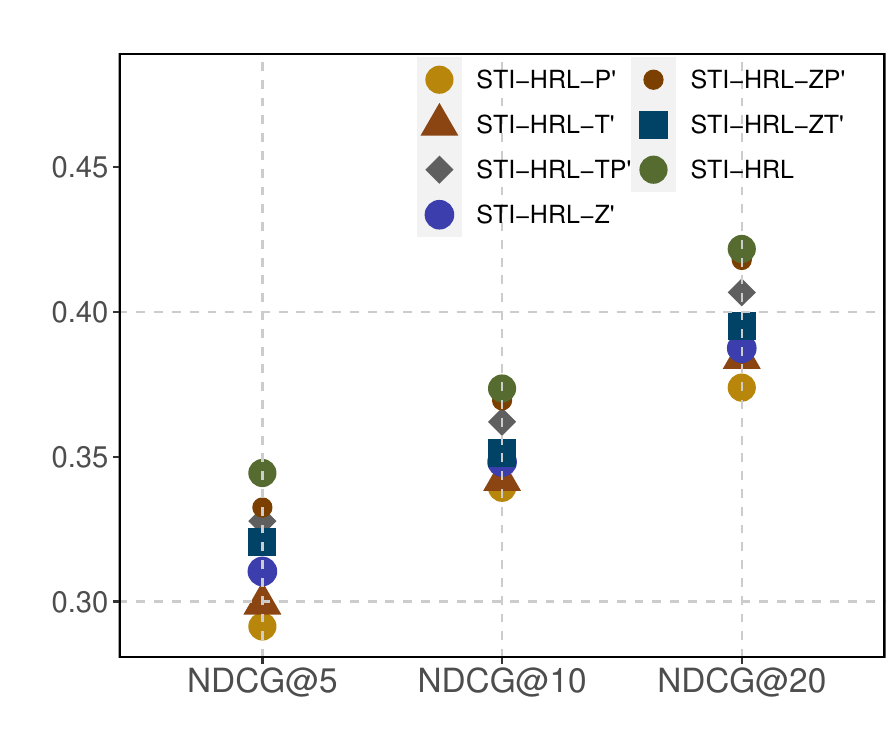}
  \scriptsize (a) New York
\end{subfigure}%
\begin{subfigure}{.24\textwidth}
  \centering
  \includegraphics[width=\linewidth]{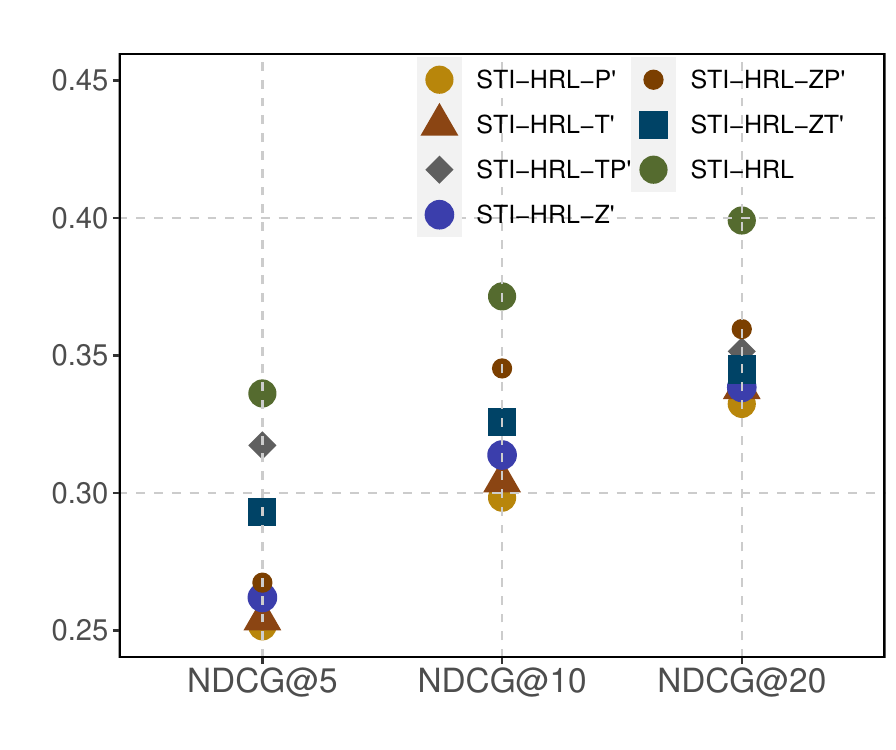}
  \scriptsize (b) Tokyo
\end{subfigure}
\caption{An ablation study on hyperedge types.}
\label{fig:ablation study on hyperedges}
\end{figure}

\subsection{The Analysis of Spatial and Temporal Factors (Q2)} 
We visualize the weights of policy integration for spatial ($\beta$) and temporal ($1-\beta$) factors in the prediction, and present the distribution in 
Figure~\ref{fig: spatial-temporal factors}. 
The results illustrate trade-off preferences towards spatial and temporal factors varies across users. 
It's evident that many users display a pronounced bias towards either time or space, suggesting these elements play pivotal roles in their decision-making. This indicates a discernible behavioral pattern among users when opting for mobility decisions.

\subsection{The Study of STI-HRL Agent Architecture (Q3)} 
To investigate the design of the two-layered agents in STI-HRL, we introduce three variants of STI-HRL:  STI-HRL-$S'$, STI-HRL-$T'$, and STI-HRL-$I'$, denoting the variants that spatial, temporal, and high-level integration removed agents, respectively. 
We compare the performances of the three variants and STI-HRL,  and present the results in~Figure~\ref{Fig: ablation study on agents}. 
The results indicate that STI-HRL performs the best, suggesting the necessity of two-layered decoupling-integration design of agents. 
Moreover, the performance of STI-HRL-$S'$ and STI-HRL-$T'$ shows that spatial and temporal factors are both essential for human mobilities, which further justifies the critical role of spatial-temporal interplay.

\begin{figure}[!th]
\centering
\begin{subfigure}{.24\textwidth}
  \centering
  \includegraphics[width=\linewidth]{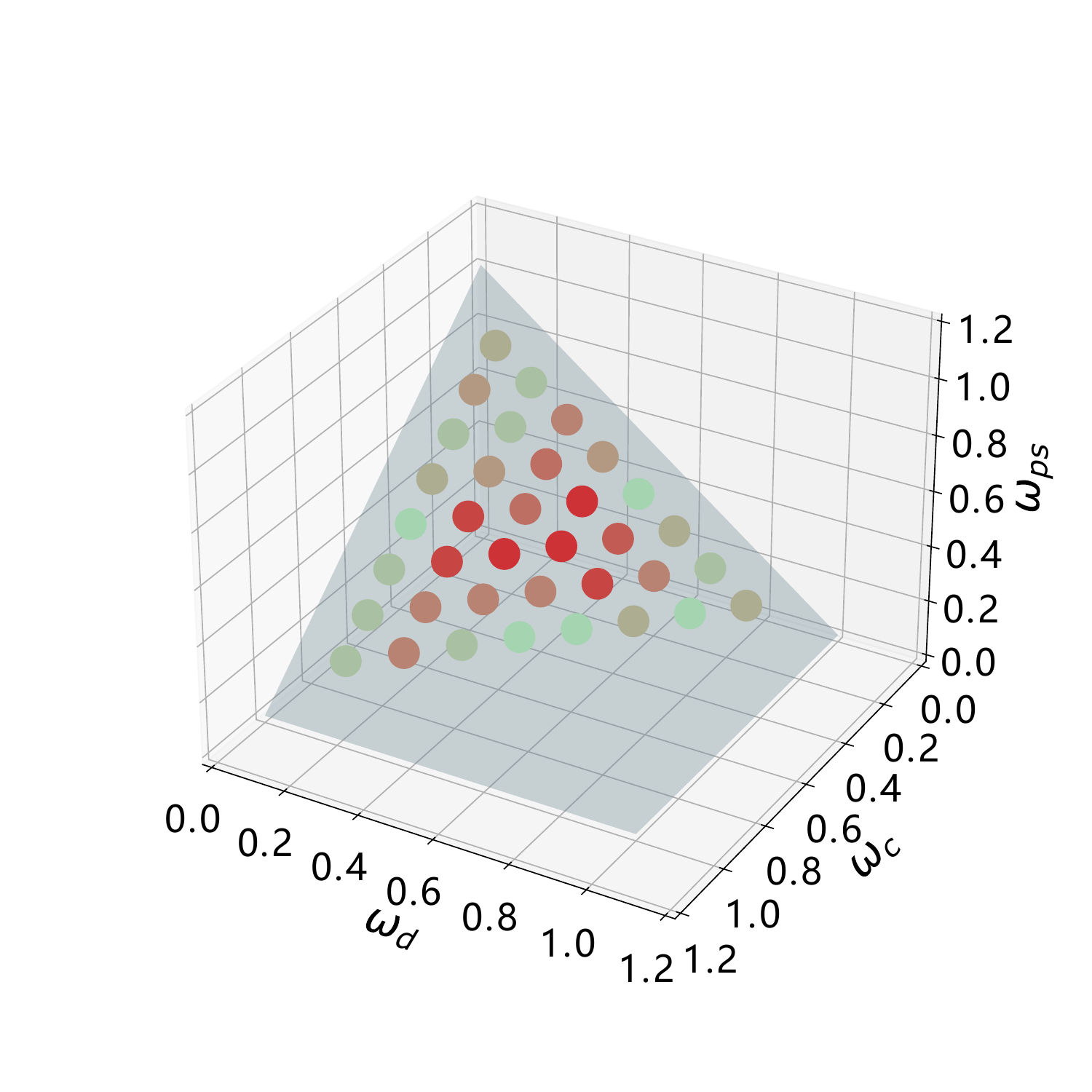}
  \scriptsize (a) New York
\end{subfigure}%
\begin{subfigure}{.24\textwidth}
  \centering
  \includegraphics[width=\linewidth]{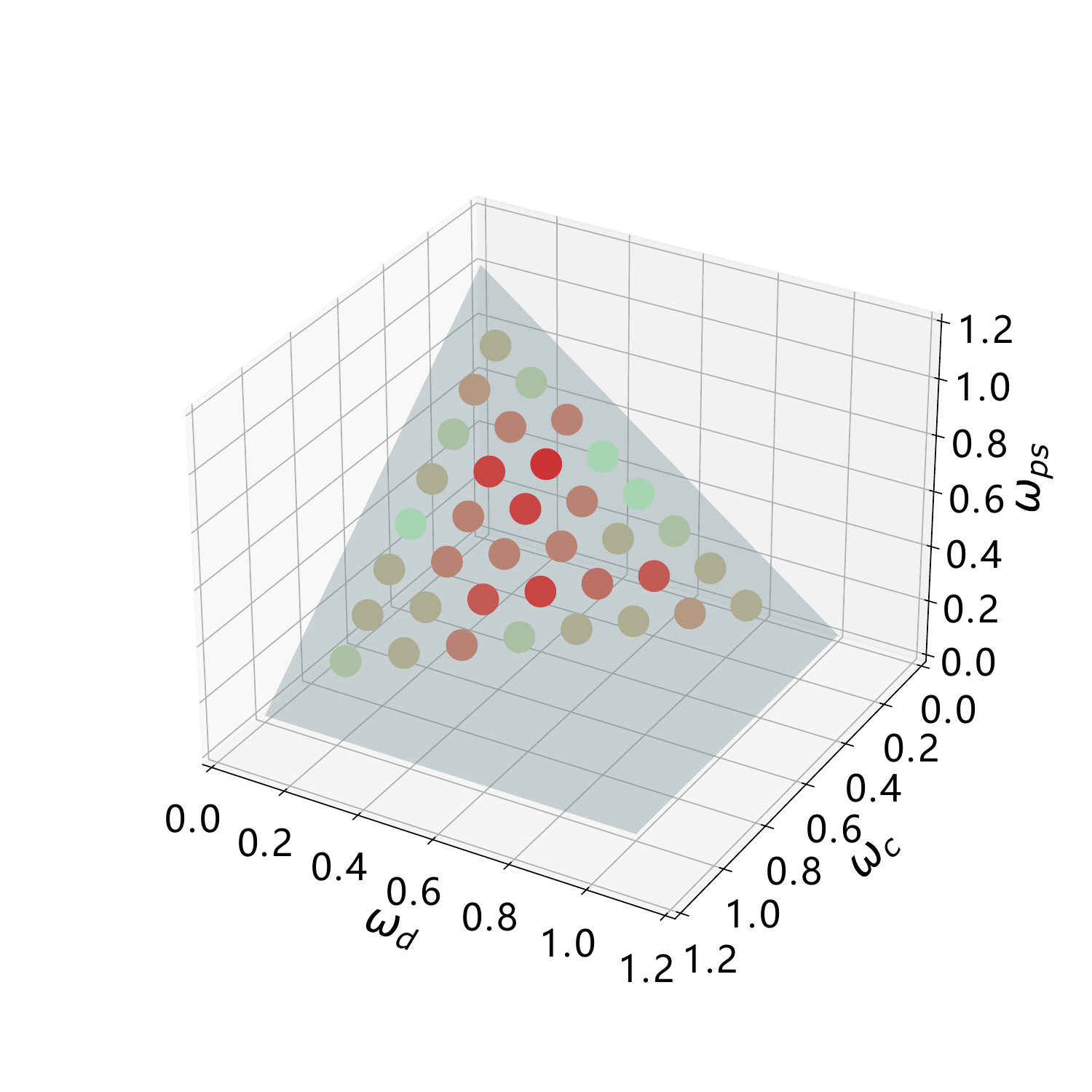}
  \scriptsize (b) Tokyo
\end{subfigure}
\caption{The analysis of reward of the spatial agent.}
\label{fig: spatial reward}
\end{figure}

\begin{figure}[!th]
\centering
\begin{subfigure}{.24\textwidth}
  \centering
  \includegraphics[width=\linewidth]{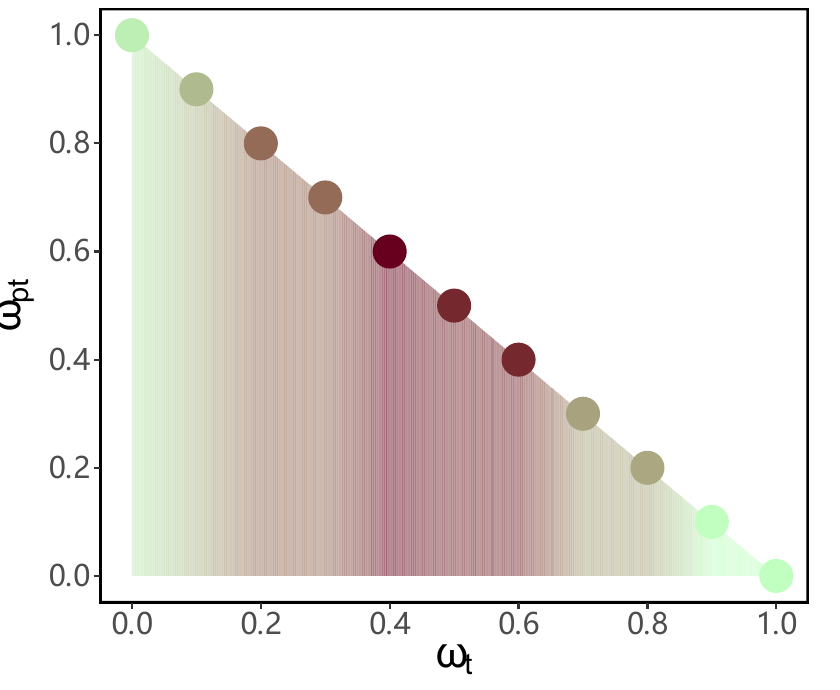}
  \scriptsize (a) New York
\end{subfigure}%
\begin{subfigure}{.24\textwidth}
  \centering
  \includegraphics[width=\linewidth]{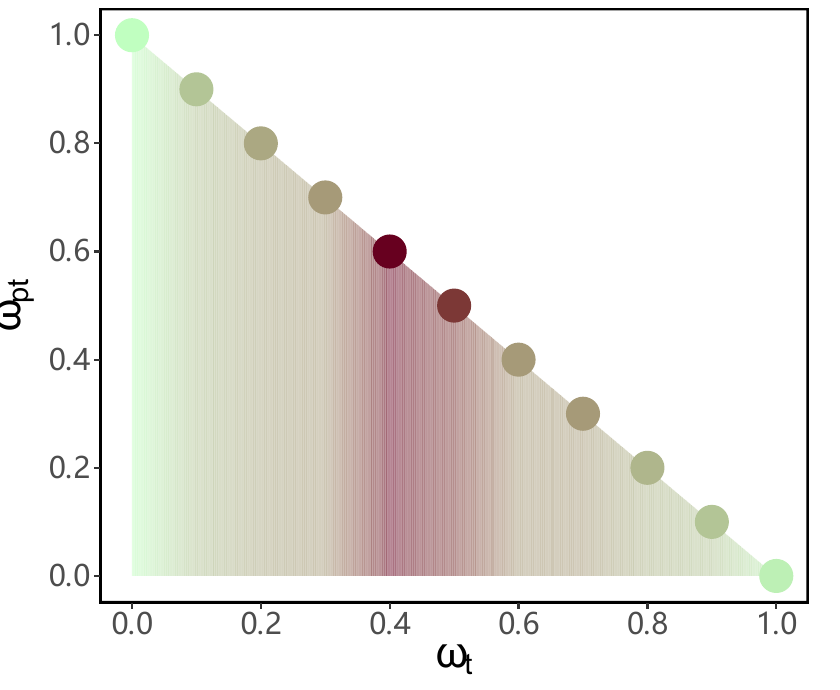}
  \scriptsize (b) Tokyo
\end{subfigure}
\caption{The analysis of reward of the temporal agent.}
\label{fig: temporal reward}
\end{figure}

\subsection{The Study of Mobility Hypergraph (Q4)}
\noindent \textbf{The design of Mobility Hypergraph.} 
We conduct an ablation experiment to verify the necessity of the hypergraph architecture. In this experiment, we design a variant of STI-HRL called STI-HRL', which directly takes the user's sequence as input without using a hypergraph architecture. As shown in Figure \ref{fig:ablation study on hypergarph}, STI-HRL shows a clear performance advantage compared to STI-HRL'. This result highlights the effectiveness of the hypergraph architecture in dealing with such problems.

\noindent\textbf{The effects of hyperedge types.}  
We take the embeddings of three types of hyperedges (POI, Time, and Zone) as the state.
Therefore, we introduce six variants of STI-HRL to study the effects of hyperedge types, including
(1) STI-HRL-$P'$, POI hyperedges only; (2) STI-HRL-$T'$, time hyperedges only; (3) STI-HRL-$Z'$, zone hyperedges only; (4) STI-HRL-$ZT'$, zone and time hyperedges; (5) STI-HRL-$TP'$, time and POI hyperedges; (6) STI-HRL-$ZP'$, zone and POI hyperedges. 
We conduct the ablation study on the six variants and present the results in Figure~\ref{fig:ablation study on hyperedges}. 
The results demonstrate that varying permutations and combinations of hyperedge types yield different performances. 
This indicates the significance of the semantics conveyed by these hyperedge types in representing human mobility events. 
Only STI-HRL, which incorporates all hyperedge types, achieves optimal performance.

\subsection{The Analysis of Reward Design (Q5)} 
We consider different combinations of weight setting of reward functions to test the performance of STI-HRL. 
Since the high-level reward settings ($w_S$ and $w_T$) are automatically learned, which cannot be manually adjusted. 
Therefore, we only analyze the reward setting of spatial and temporal agents.
Specifically, for the spatial agent reward $r_S$, three components ($r_d$, $r_c$, $r_p$) are considered. We project the performance of different combinations of $w_d$, $w_c$, and $w_{ps}$ ($w_d+w_c+w_{ps}=1$) to 3D space, and mark the performance in different scales of color, as shown in Figure~\ref{fig: spatial reward}. 
The better the performance, the darker the color. 
For the temporal agent reward $r_S$, two components ($r_t$ and $r_p$) are considered. 
We plot the results in Figure~\ref{fig: temporal reward}. 

\section{Related Work}
\subsection{Human Mobility Prediction}
Human mobility modeling centers on predicting users' next-visit locations by harnessing insights from historical check-in data. The rise of deep learning introduced advanced techniques like Recurrent Neural Networks (RNNs)~\cite{liu2016predicting,yang2020location} and self-attention mechanisms~\cite{lian2020geography,DBLP:conf/www/LuoLL21}. While they excel at modeling sequential regularities, especially with spatial-temporal data, they tend to oversimplify complex spatial-temporal interplay influencing user choices.

\subsection{Hypergraph Learning}
Compared to other graph models~\cite{dong2023temporal,DBLP:conf/aaai/00070WWY23,dong2023adaptive} that deal with heterogeneous information,
hypergraph learning~\cite{yang2019revisiting,liu2022hypergraph,yan2023spatio} offers a robust representation of the nuanced information within human mobility data, outshining traditional graph-based counterparts. Yet, even with advances in hypergraph embedding, many methods overlook daily behavioral similarities among unconnected individuals as seen in historical trajectories~\cite{yang2022getnext}, curbing the full potential of existing models.

\subsection{Reinforcement Learning}
At its core, reinforcement learning focuses on learning policy learning through environmental interactions and feedback~\cite{mnih2015human,jiang2023reinforced}. The duality of exploration and exploitation in this learning paradigm aptly captures the evolving user preferences. Hierarchical Reinforcement Learning (HRL) extends it to a wide variety of comprehensive recommendation domains ~\cite{zhang2019hierarchical,
yu2020expanrl,
xie2021hierarchical,
du2022denoising,
DBLP:journals/tmc/WangDYJMSZQW22} has a wide range of applications. Our study pioneers the application of HRL for the next visit in human mobility, approaching it from a unique disentangled viewpoint. Existing HRL models aren't directly transferrable to our challenge due to distinct goals and data structures.

\section{Conclusion}
In this study, we address the challenge of capturing the spatial-temporal interplay in human mobility. 
We frame this challenge as a two-tiered decision-making task: the low-level focuses on disentangling spatial and temporal preferences, while the high-level synthesizes these insights to produce a comprehensive mobility decision. 
Our approach leverages a hierarchical reinforcement learning framework. 
To effectively encapsulate the multifaceted semantics of spatial-temporal dynamics, we introduce a Mobility Hypergraph to structure the mobility records, utilizing hyperedge embeddings as states for policy learning. 
Comprehensive experiments validate the efficacy of our method. 
Its consistent superiority over baseline models underscores the importance of spatial-temporal interplay in human mobility modeling.

\section{Acknowledgements}
This work is supported by the Science and Technology Development Fund (FDCT), Macau SAR (File No. SKL-IOTSC-2021-2023, 0047/2022/A1, 0123/2023/RIA2), the University of Macau (File No. SRG2021-00017-IOTSC, MYRG2022-00048-IOTSC).
This work is supported by NSFC (under Grant No.61976050, 61972384, 62376048) and the Shandong Provincial Natural Science Foundation, China under Grant ZR2022LZH002.

\bibliography{aaai24}
\end{document}